\documentclass{llncs}  %
\usepackage[english]{babel}
\usepackage[T1]{fontenc}
\usepackage{graphicx}
\usepackage{url}
\usepackage{microtype}
\long\def\COMMENT#1\ENDCOMMENT{\message{(Commented text...)}\par}

\begin{document}
\title{RDF annotation of Second Life objects:}
\subtitle{Knowledge Representation meets Social Virtual reality}

\author{Carlo Bernava \and Giacomo Fiumara\inst{1} \and Dario Maggiorini\inst{2} \and Alessandro Provetti\inst{1} \and Laura Ripamonti\inst{2}}
\institute{Dept. of Mathematics and Informatics, Univ. of Messina,\\ 
V.le F. Stagno d'Alcontres, 31. I-98166, Messina, Italy.\\
\email{\{gfiumara,ale\}@unime.it}
\and
Dept. of Informatics, Univ. of Milan\\
Via Comelico 39/41, Milan. I-20135 Italy.\\
\email{\{dario, ripamonti\}@di.unimi.it}
}


\maketitle

\begin{abstract}
We have designed and implemented an application running inside Second Life that supports user annotation of graphical objects and graphical visualization of concept ontologies, thus providing a formal, machine-accessible description of objects. As a result, we offer a platform that combines the graphical knowledge representation that is expected from a MUVE artifact with the semantic structure given by the Resource Framework Description (RDF) representation of information.
\end{abstract}

\section{Introduction}
We present our project on the implementation of a platform for the creation of structured, i.e. non-flat, folksonomies and taxonomies in a MUVE (Multi User Virtual Environment).
We have designed and implemented a set of Second Life (SL)%
\footnote{Second Life is owned and developed by Linden Labs \url{http://lindenlab.com/}}
user objects (called {\em Prims}) that bring into the virtual environment i) the organization of a Folksonomy, intended as a keywords deployed user annotations.
The referred folksonomies will be located outside SL and will be represented in the standard RDF notation, thus maximizing flexibility and accessibility by non-human agents.

Our Prims allow avatars and Second Life agents from within to participate to the building of the folksonomy by contributing their own annotations.
Hence, our solution blurs the distinction between consulting/annotating objects inside and outside SL and the distinction between user- and agent-annotations.

In the landscape of contemporary Web, our project can be seen as bringing together folksonomies, a staple of Web2.0 and the SL advanced graphical visualization that is needed to participate to Multi-users graphical virtual environment.

Even though our implementation is exclusive to SL, as it relies on the Linden Lab scripting language, the methodology for annotation is not specific for SL and can be applied, in our opinion, in future collaboration platforms.

\subsection{Goals of the project}
The long-term goal of our project is the creation of a knowledge base within Second Life that is accessible both to human users (through their avatars) and SL applications (which include non-human avatars).
Our Knowledge base will support the automated access, selection and consultation of
the objects appearing on one or more specified SL loci.
The application we developed is available to all avatars in the (virtual) neighborhood of Prims (primitives of objects existing in the virtual environment); avatars will collaborate in making a semantics emerge out of frequencies and in annotating SL Prims.
By sending messages from avatars/bots or by interacting directly  with the Prims that carry the application, it is possible to insert new elements of the knowledge base so to obtain a 3D graphical representation of the relationship among concepts.

\section{Knowledge Bases and user annotations}
A Knowledge Base (KB) can be seen as a set of formal statements that express information (knowledge) about an object or a resource which is being described.
Today, almost all semantics annotations on the Web are expressed by means of RDF (Resource Description Framework) and its variations, e.g., RDFA, which allows annotations ``hidden'' in traditional HTML pages.
At a minimum, the attributes of a resource are described by means of properties which in turn are described by ontologies, which are available on the Web. The approach used in our application is named \textit{social tagging} and, as is known, produces  \textit{folksonomies}.

\subsection{Folksonomies}

Folksonomies are a popular way to classify objects through the use of tags defined by users.
It can be defined as a bottom-up classification, as opposed to the classical methods, namely taxonomies.
User annotation is one of the elements underlying the success of  social network repositories such as Flickr and YouTube.
A tag is simply a string that users \textit{attach} (associate) to the object that they are classifying, thus introducing a series of concepts somehow related to the resource.
In principle, the task of associating concepts and terms to objects is {\em continuous} and open to all users, thus resulting in a great expressivity.
With folksonomies, the frequency by which a tag is associated to a resource may become (in a social networking scenario) a statistically-accurate description of the resource itself.

Thanks to user annotation and convergence, each resource can thus be represented by means of the so-called \textit{cloud of tags}, namely a graphical representation of all tags, where a tag size represents analogically the frequency of its use in context.
Again, social tagging is meaningful only in large communities of users where the meaning of an object may {\em emerge} from the frequency by which tag terms are attached to it. 
So, even though user tagging does not provide a semantics \textit{per se,} it does provide the basis for the creation of one.

\begin{figure}[htbp]
\begin{center}
\includegraphics[scale=0.5]{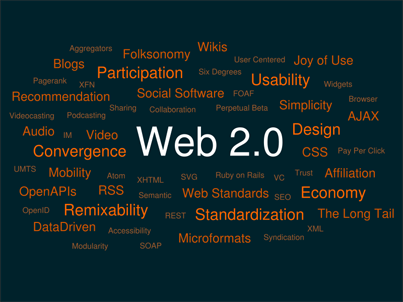}
\caption{Example of a cloud of tags created by L. Cremonini for \textit{railsonwave.com}}
\label{fig:tagCloud}
\end{center}
\end{figure}

\subsection{The SCOT ontology}\label{sec:SCOT}
In order to describe the composition of declarative languages used in our application, we have adopted the SCOT (Semantic Cloud of Tags) ontology.
As an ontology, SCOT describes in a precise way the use of various properties and classes available to developers.
We briefly recall the properties of SCOT that were used more frequently in our application.

\begin{itemize}
	\item \textbf{TagCloud}: this class describes the main repository of the tagging activity, i.e., namely the cloud of tags.
	All other properties derive from this one.

	\item \textbf{HasTag}: this property describes the set of tags associated to an object.
	It is incorporated from the \textit{sioc} ontology.

	\item \textbf{Tag}: this class describes the single tag belonging to a cloud of tags.
	It is a simple string used to identify a specified resource.

	\item \textbf{OwnAFrequency}: a RDF \textit{DataTypeProperty} used to describe the absolute frequency of occurrence of a tag, namely the total number of times the specified tag has been associated to an object.
\end{itemize}

Notice that these properties have a hierarchical structure: \textit{TagCloud} contains \textit{HasTag} which in turn contains \textit{Tag}, which contains \textit{ownAFrequency.}
In order to explain how classes and properties of the SCOT ontology are exploited to create RDF models, it is useful to see an example of a model extracted from the knowledge base of the application related to object \textit{computer mouse:}

\begin{figure}[htbp]
\begin{center}
\includegraphics[scale=.60]{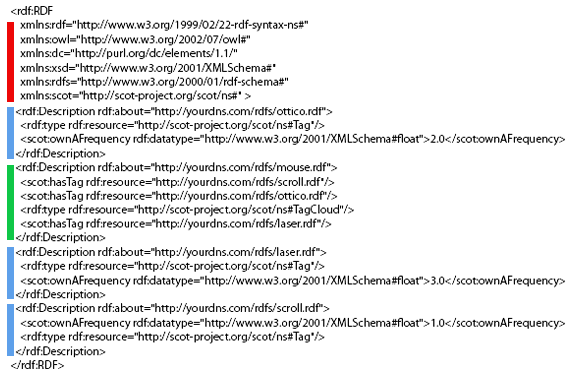}
\caption{Example of a RDF model used to describe the resource \textit{mouse}}
\label{fig:rdfMouse}
\end{center}
\end{figure}

In Figure \ref{fig:rdfMouse} an example of a RDF model is shown, describing the {\em tagging activity} of the object \textit{mouse}: \textit{scroll, optical, informatics, components, peripheral,} and \textit{laser} that has been used to describe the mouse.
For ease of reading, different parts of the model have been drawn with different colors.
Let us now see in detail what are the components of our annotations.

\begin{itemize}
	\item Preamble:
	The red section, immediately after the opening tag, is called preamble and is present in every type of model.
	It contains the namespaces used in the model. Each line specifies a namespace and its URI. It has the form:

	\begin{center}
	\texttt{xmlns:xxx=``uri ontology''}
	\end{center}

	where \texttt{xmlns} is a particular namespace associated to every RDF model and is used for dynamic binding of namespaces within the RDF model.
	The string \texttt{xmlns:xxx} specifies the prefix \texttt{xxx} which will be used for all properties belonging to the namespace specified on the right of the line.
	As an example, please note the third line in the preamble:

	\begin{center}
	\texttt{xmlns:scot=}
	\texttt{"http://scot-project.org/scot/ns"}>
	\end{center}

where the various components of the SCOT ontology will be uniquely identified using the prefix \texttt{scot} and its properties will be invoked simply typing:

	\begin{center}
	\texttt{<scot:property=value>}
	\end{center}

or, equivalently:

	\begin{center}
	\texttt{<scot:property>value</scot:property>}
	\end{center}

\item Object description: The green section relates to the description of the resource.
	Line
	
	\begin{center}
	\texttt{<rdf:type rdf:resource=}
	\texttt{"http://scot-project.org/scot/nsTagCloud"/>}
	\end{center}

describes the resource as a cloud of tags. Other lines in this section specify the structure of the cloud of tags.

\item Definition of tags:
	The blue section refers to the definition of tags and illustrates their main features.
	In this example, the \texttt{laser} tag shows an absolute frequency defined by means of a float data type.

\begin{scriptsize}
\texttt{<rdf:Description rdf:about="http://yourdns.com/rdfs/laser.rdf">\\
<rdf:type rdf:resource="http://scot-project.org/scot/nsTag"/>\\
<scot:ownAFrequency \\
rdf:datatype="http://www.w3.org/2001/XMLSchema float">3.0\\
</scot:ownAFrequency>\\
</rdf:Description>}
\end{scriptsize}
\end{itemize}

In conclusion, the SCOT ontology has been so far expressing enough.

\subsection{Taxonomies}
A taxonomy defines both a hierarchical classification of concepts and the classification methods: complex of ideas and rules with which object are classified.
The taxonomic model seems to be in contrast with the folksonomy model defined before, as in the former there is a well-founded rule to bind concepts, while in the latter there is no rule at all and the process is based on frequency.
In this application we created an hybrid model which exploits the hierarchical feature of taxonomies and the lack of rules of folksonomies.
In short, {\em we created a folksonomy organized as it were a taxonomy}.
Concepts are bound to their respective ancestors without any pre-defined rule of derivation, but are organized according to topic maps, in the form of trees.
Of course, the scalability of this model is somewhat limited, in that the tagging of all conceivable concepts would produce a graph rather than a tree.
Nevertheless, the goal of our model is the creation of a topic map, with an arbitrary depth, derived from a single concept.

\section{Second Life}
In on-line multiplayer games (the so-called MMORPG, Massively Multiplayer On-line Role-Playing Game) the user/player usually pursues objectives preset by the administrator/designer of the game and which can be changed according to feedback of players.
In Second Life, on the opposite, the user is able to use her avatar (her alter-ego in the virtual reality) as a synthetic human being, according to an evolution directly maintained and sustained by the users themselves.
This is why Second Life is usually defined a MUVE (Multi-User Virtual Environment).
In this section, we provide some explanations about terms and concepts typical of SL.

\subsection{Regions}
The world of Second Life is made up of regions, which are reachable from within the MUVE by means of specific coordinates.
Due to the intrinsically tri-dimensional nature of SL, each point is located by means of a set of four coordinates, namely island name, abscissa, ordinate and altitude.
A region is defined as a square with edges measuring 256 meters each. Also, the region contains the space over itself, up to a maximum altitude of about 166 meters.
There exist three types of regions:

\begin{itemize}
	\item mainland: large amounts of contiguous land owned by Linden Labs.

	\item private region: islands mainly inhabited by users who buy lands. The administrator of this type of regions is able to access all administrative controls. This is the most common type of region in SL, and

	\item openspace: a land with intermediate features between the former two, with limitations on the number of objects and avatars allowed to visit it.
\end{itemize}

SL allows region administrators to create groups associated to resident users in order to allow/forbid some activities such as creating objects, modifying the land or entering some zones.
Together, the regions form what is called the \textit{grid} of Second Life.
At infrastructure level each region is hosted inside a software virtual machine taking care to coordinate interaction between avatar and objects and to simulate physics.
Typically, each virtual machine is located on an hardware server whose computational power is in the same range as a mid-range desktop PC.
The legacy software manages the lands together with the interactions among avatars and the creation/modification of the objects acting in each square.

\subsection{Objects}
With SL, users have the possibility to create custom objects with a virtual physical appearance.
In order to do this, the avatar of a user must have the necessary rights, some exceptions being possible in particular regions called \textit{sandboxes}.
As for regions, the creator of an object is the owner and is allowed to change its features, including, e.g., the texture, the consistency of the material, and the color.
The creator may also forbid changes from other users.
The creation of SL objects is greatly helped by a 3D modeling tool, which supports complex operations such as torsion, dilatation and compression, concavity and convexity, flexibility.
Objects may be associated to certain types:
\begin{itemize}
	\item \textbf{locked}, in order to prevent modifications of the object;

	\item \textbf{physical}, if the object has to undergo physical laws of the emulator, such as gravity or friction;

	\item \textbf{temporary}, if the object has a limited life, a

	\item \textbf{phantom}, if other objects or avatars may pass through the created object.
\end{itemize}

Automatically, the response to physical laws such as friction depends on the matter with which objects are ``made'' of.
So, for example a glassy object has a lower friction than a woody one.

\subsubsection{Primitives and the LindenLab Scripting Language (LSL)}

Primitives (Prims) are the basic building blocks of the SL virtual environment.
Prims appear as simple geometrical objects (cones, cylinders etc.); by combining them complex objects can be created.
A complex graphical object may be composed of up to 255 Prims.
However, an avatar may create complex objects only by using Prims that he/she owns. 
All objects and avatars have a 128-bit identifying code called UUID (Universal Unique IDentifier), which is frequently refereed to within the native LindenLab scripting language, which is presented next. 

In Second Life objects may have associated scripts, which make it possible to objects to perform actions of various kind.
The native underlying scripting language is called Linden Scripting Language (LSL), a C-like, event-driven programming language.
The LSL compiler produces a bytecode which can be executed by SL servers.
Although it is endowed by a set of more than 300 functions, LSL has some important limitations:

\begin{itemize}
	\item it is loosely typed

	\item a maximum of 64KB of memory can be assigned to a script associated to a prim
\end{itemize}

\noindent which obviously rule out the possibility to endow objects with their own knowledge representation capabilities.

Other major limitations are:

\begin{itemize}
	\item money (the well-known Linden dollars) cannot be withdrawn from an avatar, unless explicitly allowed

	\item it is impossible to instruct an avatar to perform some animation, with the exception of some postures

\end{itemize}

\noindent Interaction between avatars and objects is governed by a messaging system, which can be local
to a SIM or global  to  the grid.
Local  interaction is  initiated by  the  interface (e.g., by mouse clicks or a keyboard press) and by text messages (or with the chat).
When one of these events is triggered, the simulator will distribute a number of messages  to  involved avatars  and  objects;  reception  of  these messages may imply  the  visualization  of  a  text message and/or a state change for a program inside an object.
Global interaction is essentially text-based and is mainly intended as an inter-SIM instant messaging system; both avatar and objects can benefit from global communications.
Communication is a key point of SL: the messaging  system  can  be  intertwined with  all  other in-world operations.
Also, interaction between avatars will extend in-world with no distance boundaries and will also span off to real life, because messages will be relayed via e-mail when the user is not on-line.

Expressing personal identity and feelings can be performed not just by avatar reshaping, but also by wearing (attaching) scripted Prims, which in turn will be able to interact and send messages to nearby objects and avatars.
Thus, attachments plays a role in how the surrounding environment perceives
the user presence, even at a distance.

SL places can be filled with interacting and active scripted objects, which will send messages to users independently from their whereabouts, thus, helping  creating  a  social network without the constraint of {\em being there}.
To  some  extent,  SL  communication  is  not limited  to  the  grid  itself:  scripted  objects  have means  to  reach  the  Internet and use data  from  it to augment the virtual environment; importantly, it is possible to access web content as well as send/receive e-mail messages and even SMS (Short Message System).
Multimedia  content  from  the  Internet  is  supported in an indirect way; a real-time media stream can be set as part of the environment. Thus, the SL client will take care to independently retrieve the content and perform the playout without interfering with the grid.

\section{Architecture of the application}
SL has a rich set of features, but communications with
external applications and the scripting of Prims are two weak points, especially in view of the deployment of MASs.
Indeed, scripts are limited to a memory occupation of about 25 KB, which includes the source code itself
and the prim description (name, owner, texture, interactions prim-prim and avatars-prim, permissions etc.).
Also, communication with external applications may occur only throughout XML-RPC and HTTP.
The former is not suited for intensive applications as a consequence of the delay artificially introduced between two subsequent requests.
The latter is also severely limited to about 25 maximum requests over 20 seconds, preferably with the GET method, and to a total amount of 2048 bytes each.

\begin{figure}[htbp]
\begin{center}
\includegraphics[scale=0.2]{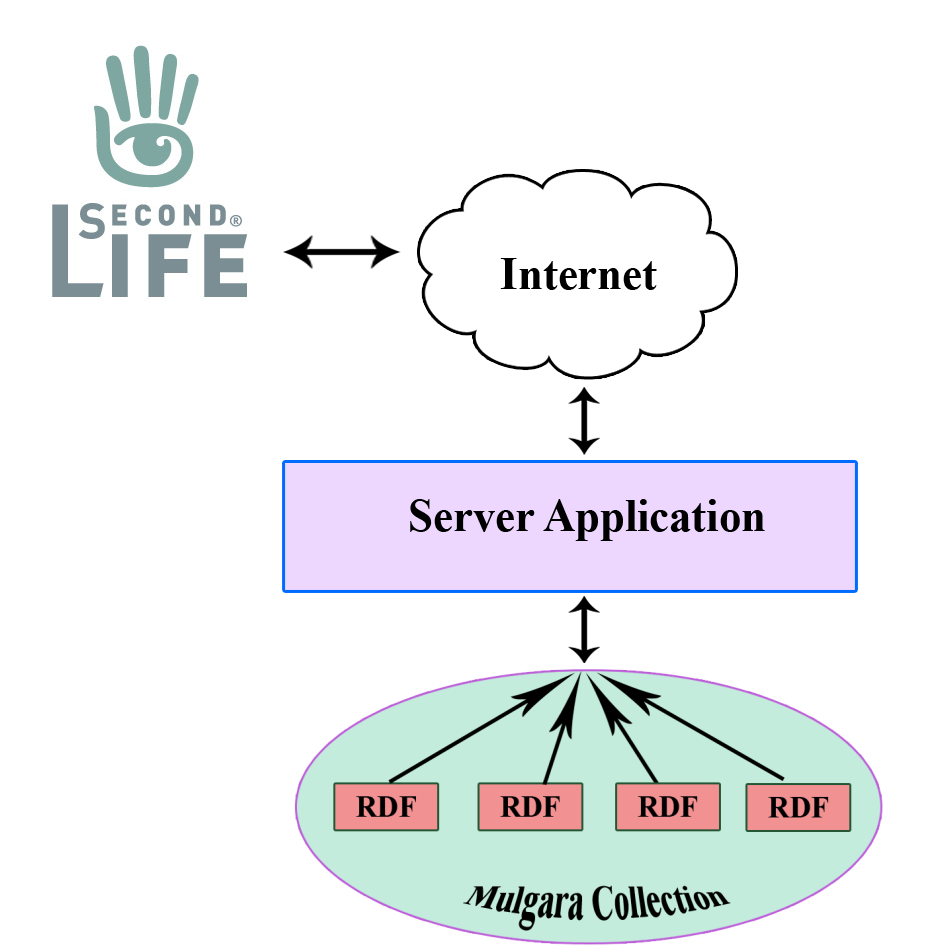}
\caption{Architecture of the application}
\label{fig:arch}
\end{center}
\end{figure}

\begin{figure}[htbp]
\begin{center}
\includegraphics[scale=0.2]{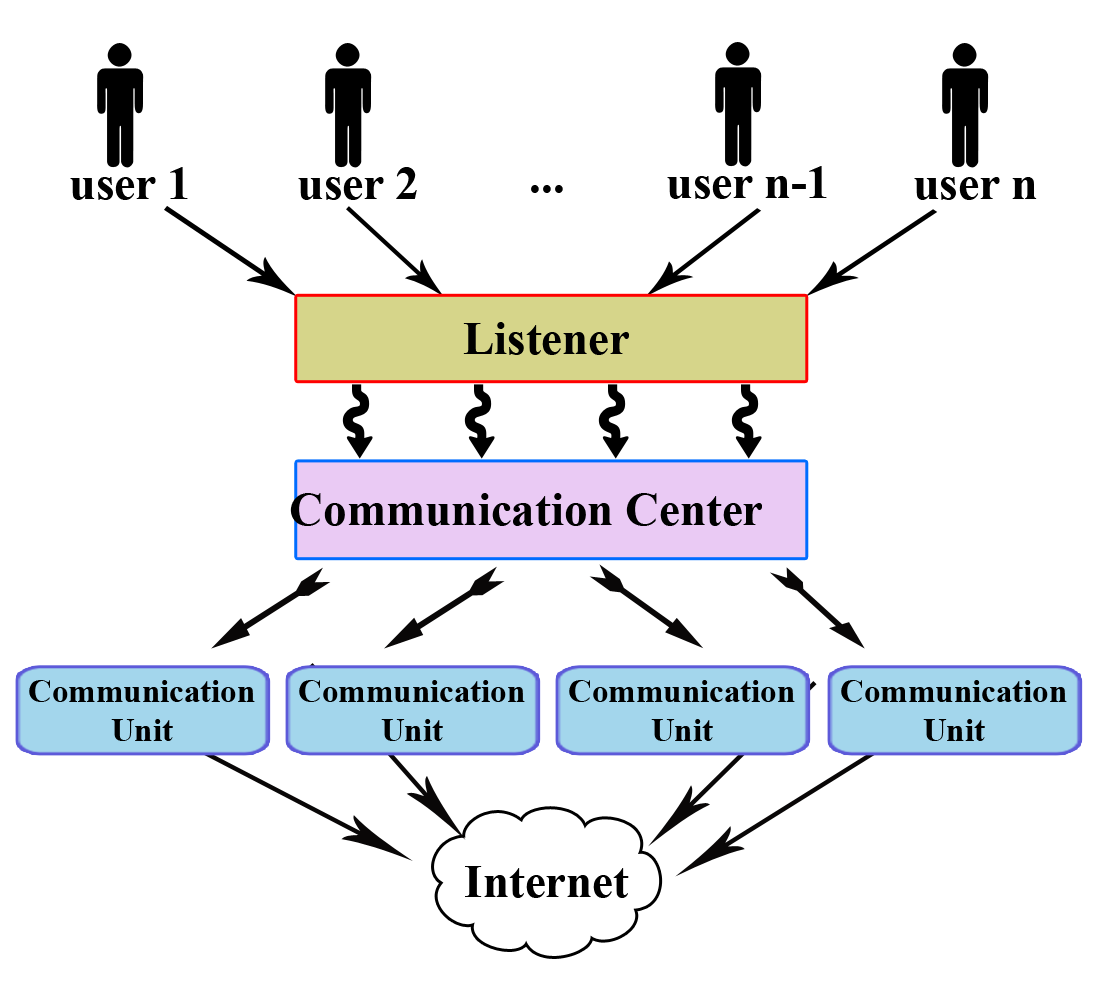}
\caption{The architecture of Second Life client-side of our application}
\label{fig:arch-detail}
\end{center}
\end{figure}

The main implementation effort of our project so far has been the development of an architecture based on the client-server paradigm: clients from within SL (see Figure \ref{fig:arch}) communicate with on an external server by means of HTTP requests.
In order to overcome these potential bottlenecks we resorted to implement a light application on the SL client-side (see Figure \ref{fig:arch-detail}).
Our solution consists of a listener unit which is activated by avatars; in turn it activates a ``communication center'' from which a communication unit is originated for each request.
Our solution bypasses both the limitations on the maximum number of requests and on the maximum script size.
Communications among the various scripts are made possible using the channels that within Second Life are provided for conversations among avatars and/or scripts.
The server side of the application is composed of a Java servlet that

\begin{enumerate}
	\item receives requests from clients;
	\item exploits the SCOT ontology (described in Section \ref{sec:SCOT}) to produce RDF models;
	\item serializes them into XML-RDF structures to be stored in a Mulgara RDF database (see the example in Figure \ref{fig:rdfMouse}).
\end{enumerate}

The graphical rendering of ontologies in SL has been relatively easy in comparison.
However, in SL there is a limitation on the maximum distance, about 10 meters, at which a Prim can be {\sl rezzed} by a script.
Such limitation implies a strict constraint on the maximum extension of the graphical appearance of the concept map.
As the size of the hierarchy grows, a fixed maximum space occupation may affect negatively (human) readability, yet we have not yet experienced such degradation with fairly complex conceptual maps.

\section{Results}
\begin{figure}[htbp]
\begin{center}
\includegraphics[scale=0.4]{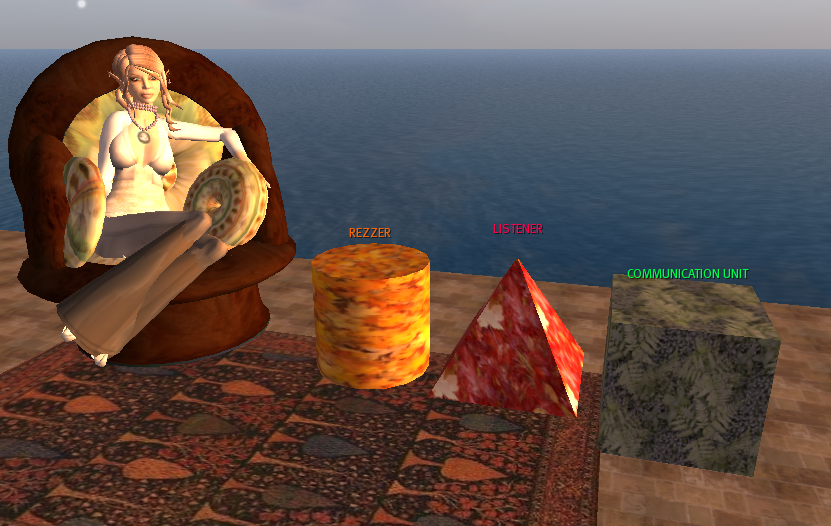}
\caption{An avatar sees our Prims}
\label{fig:prims}
\end{center}
\end{figure}

The main technical result of our project is the first-time embedding of a knowledge base with a graphical, interactive environment of the size and complexity of Second Life.
The chosen application can be sketched as in Figure \ref{fig:prims}.

The Prims showed in Figure \ref{fig:prims} contain the client-side application.
They have been left as primitive as possible from the graphical point of view, so land owners are free to customize their graphical part.
Moreover, as the application has been conceived for the use of bots, its graphical simplicity makes easier the interaction with the application.

\begin{figure}[htbp]
\begin{center}
\includegraphics[scale=0.3]{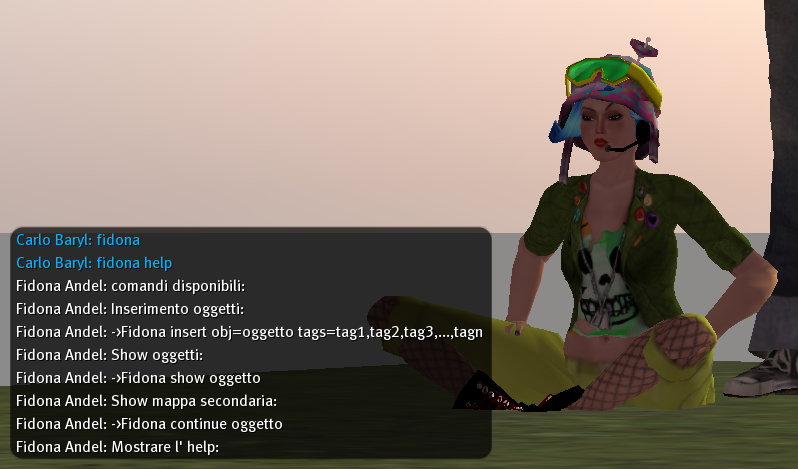}
\caption{Submission of a query to the bot}
\label{fig:terzo}
\end{center}
\end{figure}

\begin{figure}[htbp]
\begin{center}
\includegraphics[scale=0.2]{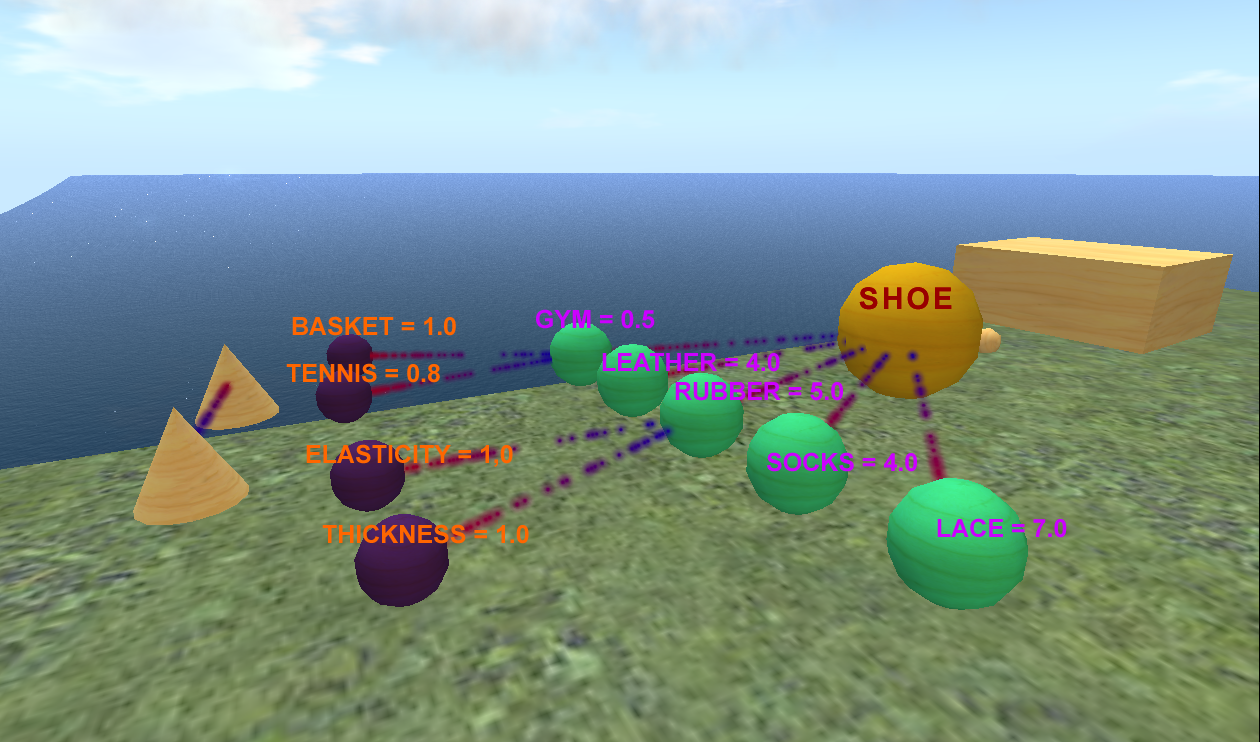}
\caption{Example of a two-level topics map for resource \textit{shoe}}
\label{fig:2-levels-map}
\end{center}
\end{figure}

As can be seen in Figure \ref{fig:terzo}, the avatar is querying the bot by asking which messages activate the software components.
As shown in Figure \ref{fig:2-levels-map}, each tag associated to the resource is annotated with its absolute frequency.
With the same tools, multi-level topic maps may be built.
In Figure \ref{fig:2-levels-map}, a two-level topic map is shown.
The first level is developed starting from the concept of \textit{shoe}, while the second level details \textit{rubber,} the material this shoe sole is made of.

Topic maps can cover great lengths (w.r.t. Second Life virtual reality), so the client application must be viewed in locations free from other objects that could interfere with the proper visualization.
In order to preserve performances in the space which hosts the maps, and to avoid problems due to residual maps, topic maps undergo an automatic self-destruction after 1 minute.

Estimating the performances of the application is not easy, as it depends on various features.
The main factors affecting responsiveness are:

\begin{itemize}
	\item Degree of network congestion;
	\item load at server hosting our application;
	\item number of Prims active on the island considered, and
	\item complexity of the submitted query.
\end{itemize}

Altering any of these parameters results in a performance degradation proportional ---empirically--- to the degree of change of the parameter.

It is important to notice that the knowledge base (the set of annotations) resides in a RDF native repository, and is reachable by the application within Second Life and/or HTTP protocols and/or plain TCP/IP connections.
Our multi-protocol database allows the content usable in different contexts, so assuring the stability and availability of data from within different application, each developed in environments external to Second Life.
Even though the adopted software architecture is somehow forced by the limitations of the SL platform, for our purposes it is interesting as it brings about the benefits of having [practically all] data stored independently, and externally to the applications, even of the SL platform itself.
The obvious upshot is a certain complexity in the implementation of external calls, and their number may affect scalability eventually.
These aspects are discussed in depth by Campi et al. \cite{CamGotHoy09}.

\section{Related works}
Since its appearance Second Life attracted a lot of attention from users and several research efforts have aimed at coupling virtual environment with semantic web applications running on web servers external to SL.
Karakatsiotis et \cite{KGA08,KLA09} developed a {\em robotic avatar} (perhaps better termed a {\em bot)} which follows a virtual visitor in her tour of the virtual museum.
Whenever the visitor touches an exhibit the bot produces a textual description developed in NaturalOwl, a Natural Language Generation engine developed in Java.
A more scalable version of this work is described in \cite{4219770}.

A somewhat different perspective was pursued by Kleinermann et al. \cite{KDeTCP07}, who designed a tool, based on Ajax3D and that was mainly intended for X3D applications, for the semantic annotation and creation of navigation paths in virtual environments.
Specific to the Second Life virtual environment is the work of Schneider et al. \cite{semSL08}.
Their approach allows users to tag her own virtual objects.
It is also possible to associate a piece of software, a so-called \textit{tagget}, to a virtual object thus allowing other users to freely tag it.

Specific to the SL virtual environment is the work by Schneider et al. \cite{semSL08}: their approach allows users to tag their own virtual objects; it is also possible to associate a piece of software, a so-called \textit{tagget,} to a virtual object so to allow other users to tag it.

We would like to stress, however, that none of those works focuses on the problem of including social interaction in the process of representing and sharing knowledge.

\section{From annotation to autonomous agent systems}
As virtual environments become increasingly open, an emerging research issue is the introduction of automated external agents.
Two recent works by Rehm and Rosina \cite{RehRos08} and by Sierra et al. \cite{BRSCS09} show that Second Life may eventually embed MAS applications in full. 
Rehm and Rosina, for instance, developed an application that turns Second Life into a realistic test range for MAS; it supplies a set of features that are interesting for testing agent's social interaction skill and general long-term behavior.
Their work is a good example of how and why MAS research may adopt Second Life as a \textit{natural} setting for experiments and deployment.

Sierra et al., vice versa, show the huge potential for applying MAS to improving the Second Life experience.
Their work propose Normative Multi-agent systems as the basis for creating with a MUVE realistic, individual-centered virtual heritage applications, where users may experience a version of \textit{being there} related to ancient times, yet preserving the full range of complex individual behaviors and interactions.

Probably the MAS approach that best fits our scenario is that based on the Knowledge representation language DALI \cite{jelia02,jelia04}, which endows agents with automated reasoning, meta-reasoning and reflective capabilities \cite{bcdl:jlc00}. 
Another feature of DALI agent that we feel needed for SL agents is the explicit representation and management of complex preferences \cite{CosCommonsense07,CostantiniD12}. 
In our opinion, the DALI deployment that best illustrate this potential is \cite{CMTT08}, where agents assist the user in a visit to a museum or archeological site while taking into account her/his preferences, background knowledge,experience, time for the visit, and proposes a personalized route augmented with suggestions and information so as to lead to an enriched experience. 

\section{Discussion: what else can be done in Second Life?}
Today, there are Semantic Web, Web2.0 and MUVE applications that are very effective in their own
domain of application.
However, it is not always easy to combine the good features of each one of these domains into one.
The problem is partly conceptual and mostly technological, e.g., due to the limitations of the SL virtual environment described earlier in this article.
Ripamonti et al. \cite{RipPer10} have summarized the current situation with a use-cases table which we report, with some adaptations, in Table \ref{tab:RipPer10}.

\begin{table}[htb]
{\tiny
		\begin{tabular}{|l|l|l|}
			\hline
			  Requirements & Compl. & Notes\\
			  \hline
			 creation/representation of 3D multi-level conceptual maps & average & achievable through SL building facilities\\
			&& plus LSL scripting language;\\
			  && to guarantee a good user experience must\\
			  && -obey strict constraints intrinsic to client app.\\
			  && -address latency due to bandwidth etc.\\
			  \hline
			update of tags attached to concepts & average & \\
			\hline
			import/export maps from/to Internet sources & average & \\
			\hline
			multimedia content attach. to concepts & hard & \\
			\hline
			manipulation of multimedia content attached to concepts & hard & \\
			\hline
			definition of groups of users & easy & SL client allows groups and permissions manag.\\
			\hline
			work collaboratively on the conceptual maps & easy & collaborative work is a key feature\\
			&& yet bandwidth, server/clients \\
			&& limitations may harm users' experience\\
			\hline
			personalization of the application & average & under appropriate permissions for both code\\
			&& \& 3D it is possible (and easy) to modify objects\\
			&& \& conceptual maps appearance\\
			\hline
			definition of privacy policies to protect intell. property & easy & restricted areas are supported; SL has\\
			 && policies to guarantee IPRs of in-world artifacts\\
			 && Creative Commons licenses are available \\
			\hline
			users learn easily how to interact in the environment & easy & basic interaction is almost effortless\\
			 && 3D-modeling requires learning curve\\
			\hline
		\end{tabular}
	\caption{What a in-world application for conceptual maps should provide}
	\label{tab:RipPer10}
}
\end{table}

\section{Discussion:  how these results impact organizations}
Barriers against the integration of different organizational functions are a well-known phenomenon, whose complexity (and related costs) grows with the degree of virtualisation of the organization, reaching their maximum in the so-called Virtual Enterprises (VEs) \cite{DawMal92,GolNagPre95,Gor03}.

Several well-known case studies (see, e.g., \cite{AguMon11,RipPer10}) offer us intriguing test beds for verifying to what extent social interaction in organizational settings can be effectively supported by a MUVE.
Consider, in particular the following cases.
CIGNA-Vielife (which considers SL a key factor to communicate in an effective way to audiences that are both diverse and widespread: an environment that is both immersive and synthetic greatly facilitates lowering social taboos about sensitive topics and overcoming geographical barriers. 
For instance, IBM established one of the most complex and advanced corporate presences in SL, which lasted from 2006 to mid 2011. 
Other examples are Intel (which also exploited SL as an accessible, low-cost and carbon-free location to hold its 2009 biannual conference), US National Oceanic and Atmospheric Administration, NOAA and Northrop Grumman Corporation (NGC), who demonstrated that MUVEs seem to be an elective channel for fostering undergoing social interaction, communication and knowledge sharing among diverse and geographically distributed nets of people/communities of practice. 

In other words, SL interactions may indeed provide support to \textit{tacit knowledge sharing} \cite{Pol67,Pru01} and to the higher levels of the \textit{semiotic ladder} \cite{Sta96} in organizational settings.
Moreover, activities held in a specific virtual place create memories that could be useful when a similar situation arises in the physical world, as supported by scientific and empirical evidences in the field of game design (see e.g. \cite{Kos05} and, e.g., in the \textit{Chocolate factory} experiment held in Multiverse and described in \cite{BKRDLGFSV10}. 
In summary, we submit that virtual worlds equipped with knowledge representation should be regarded at as suitable digital habitats, able to support the higher levels of the semiotic ladder, and thus as useful tools, for example, to help lowering inter-functional or inter-team(s) barriers in organizational settings.

\section{Future work}
The architecture presented in this article represents a bridge between two different technologies, namely the Semantic web and MUVEs like Second Life.
From this point of view, our application may represent a starting point for communities accessible inside and outside Second Life.
One aspect that should be developed further is an hierarchical permission structure that
grants appropriate, role-based permissions to avatars willing to annotate objects.

A natural development of the ideas underlying our application is the possibility of porting inside Second Life Web communities which mainly based their activities on social tagging, folksonomies and taxonomies.
Also, Wikis, the Blogosphere and social bookmarking portals should \textit{export} their communities and activities.
Another possible development of our application is the implementation of reasoning techniques to data structured in a semantic way.

\bibliography{../knowrep-in-secondlife}

\begin{thebibliography}{10}
\providecommand{\url}[1]{{#1}}
\providecommand{\urlprefix}{URL }
\expandafter\ifx\csname urlstyle\endcsname\relax
  \providecommand{\doi}[1]{DOI~\discretionary{}{}{}#1}\else
  \providecommand{\doi}{DOI~\discretionary{}{}{}\begingroup
  \urlstyle{rm}\Url}\fi

\bibitem{AguMon11}
Aguiar, S., Monte, P.: Virtual worlds for c2 design, analysis, and
  experimentation.
\newblock In: Proc. of the 16th Int'l Command and Control Research and
  Technology Symposium Symposium (2011).
\newblock \urlprefix\url{http://www.dtic.mil/dtic/tr/fulltext/u2/a547157.pdf}

\bibitem{BKRDLGFSV10}
Back, M., Kimber, D., Rieffel, E.G., Dunnigan, A., Liew, B., Gattepally, S.,
  Foote, J., Shingu, J., Vaughan, J.: The virtual chocolate factory: mixed
  reality industrial collaboration and control.
\newblock In: A.D. Bimbo, S.F. Chang, A.W.M. Smeulders (eds.) ACM Multimedia,
  pp. 1505--1506. ACM (2010)

\bibitem{bcdl:jlc00}
Barklund, J., Dell'Acqua, P., Costantini, S., Lanzarone, G.A.: Reflection
  principles in computational logic.
\newblock J. of Logic and Computation \textbf{10}(6), 743--786 (2000)

\bibitem{BRSCS09}
Bogdanovych, A., Rodriguez, J.A., Simoff, S.J., Sierra, A.C.C.: Developing
  virtual heritage applications as normative multiagent systems.
\newblock In: AOSE, Lecture Notes in Computer Science. Springer (2009)

\bibitem{CamGotHoy09}
Campi, A., Gottlob, G., Hoye, B.: Wormholes of communication: Interfacing
  virtual worlds and the real world.
\newblock In: I.~Awan, M.~Younas, T.~Hara, A.~Durresi (eds.) AINA, pp. 2--9.
  IEEE Computer Society (2009)

\bibitem{CosCommonsense07}
Costantini, S., Dell'Acqua, P., Tocchio, A.: Expressing preferences
  declaratively in logic-based agent languages.
\newblock In: Proc. of Commonsense'07, the 8th International Symposium on
  Logical Formalizations of Commonsense Reasoning, AAAI Spring Symposium Series
  (2007)

\bibitem{CostantiniD12}
Costantini, S., Gasperis, G.D.: Complex reactivity with preferences in
  rule-based agents.
\newblock In: A.~Bikakis, A.~Giurca (eds.) Rules on the Web: Research and
  Applications, RuleML 2012 - Europe, Montpellier, France, August 27-29, 2012.
  Proceedings, \emph{Lecture Notes in Computer Science}, vol. 6826, pp.
  167--181. Springer (2012)

\bibitem{CMTT08}
Costantini, S., Mostarda, L., Tocchio, A., Tsintza, P.: Dalica agents applied
  to a cultural heritage scenario.
\newblock IEEE Intelligent Systems, Special Issue on Ambient Intelligence
  \textbf{23}(8) (2008)

\bibitem{jelia02}
Costantini, S., Tocchio, A.: A logic programming language for multi-agent
  systems.
\newblock In: Logics in Artificial Intelligence, Proc. of the 8th European
  Conf., JELIA 2002, LNAI 2424. Springer-Verlag, Berlin (2002)

\bibitem{jelia04}
Costantini, S., Tocchio, A.: The {DALI} logic programming agent-oriented
  language.
\newblock In: Logics in Artificial Intelligence, Proc. of the 9th European
  Conf., JELIA 2004, LNAI 3229. Springer-Verlag, Berlin (2004)

\bibitem{DawMal92}
Davidow, W., Malone, M.: The virtual corporation: structuring and revitalizing
  the corporation for the 21st century.
\newblock Harper Collins (1992)

\bibitem{KGA08}
Galanis, D., Karakatsiotis, G., Lampouras, G., Androutsopoulos, I.: Naturalowl:
  Generating texts from owl ontologies in protege and in second life.
\newblock In: 18th European Conference on Artificial Intelligence (2008)

\bibitem{KLA09}
Galanis, D., Karakatsiotis, G., Lampouras, G., Androutsopoulos, I.: An
  open-source natural language generator for owl ontologies and its use in
  protege and second life.
\newblock In: EACL (Demos), pp. 17--20. The Association for Computer
  Linguistics (2009)

\bibitem{GolNagPre95}
Goldman, S., Nagel, R., Preiss, K.: Agile competitors and virtual
  organizations.
\newblock Van Nostrand Reinhold (1995)

\bibitem{Gor03}
Goranson, T.: Architectural support for the advanced virtual enterprise.
\newblock Computers in Industry \textbf{51}, 113–--125 (2003)

\bibitem{KDeTCP07}
Kleinermann, F., Troyer, O.D., Creelle, C., Pellens, B.: Adding semantic
  annotations, navigation paths and tour guides to existing virtual
  environments.
\newblock In: T.G. Wyeld, S.~Kenderdine, M.~Docherty (eds.) VSMM, \emph{Lecture
  Notes in Computer Science}, vol. 4820, pp. 100--111. Springer (2007)

\bibitem{Kos05}
Koster, R.: A theory of fun for game design.
\newblock Paraglyph Press (2005)

\bibitem{4219770}
Oberlander, J., Karakatsiotis, G., Isard, A., Androutsopoulos, I.: Building an
  adaptive museum gallery in {S}econd {L}ife.
\newblock In: Proc. of Museums and the Web 2008 (2008).
\newblock \urlprefix\url{http://nlp.cs.aueb.gr/pubs/mw2008\_preprint.pdf}

\bibitem{Pol67}
Polanyi, M.: The Tacit dimension.
\newblock Routledge (1967)

\bibitem{Pru01}
Prusak, L.: Where did the knowledge management come from?
\newblock IBM Systems Journal \textbf{40}, 1002--1007 (2001)

\bibitem{RehRos08}
Rehm, M., Rosina, P.: Secondlife as an evaluation platform for multiagent
  systems featuring social interactions.
\newblock In: AAMAS (Demos), pp. 1663--1664. IFAAMAS (2008)

\bibitem{RipPer10}
Ripamonti, L.A., Peraboni, C.: Managing the design-manufacturing interface is
  {VEs} through {MUVEs:} a perspective approach.
\newblock International Journal of Computer Integrated Manufacturing, Special
  issue on Semiotics-based Manufacturing Systems Integration \textbf{23}(8--9),
  758--776 (2010).
\newblock DOI:10.1080/09511921003682630

\bibitem{semSL08}
Schneider, M., Kratzer, F., Mainzer, K.: semsl: Tagging and data linking for
  second life.
\newblock In: 7th International Semantic Web Conference (ISWC2008) (2008).
\newblock
  \urlprefix\url{http://data.semanticweb.org/conference/iswc/2008/paper/poster_demo/76}

\bibitem{Sta96}
Stamper, R.: Signs, information, norms and systems. Signs at work.
\newblock De Gruyter (1967)

\end{thebibliography}

\end{document}